\documentclass{article}

\usepackage{arxiv}

\usepackage[utf8]{inputenc} 
\usepackage[T1]{fontenc}    
\usepackage{hyperref}       
\usepackage{url}            
\usepackage{booktabs}       
\usepackage{amsfonts}       
\usepackage{nicefrac}       
\usepackage{microtype}      
\usepackage{lipsum}
\usepackage{graphicx}
\graphicspath{ {./images/} }

\def\eg{\emph{e.g.}}
\def\ie{\emph{i.e.}}

\def\wrt{\emph{w.r.t. }}

\def\name{StyleTailor}
\newcommand{\bfsection}[1]{\vspace*{0.0mm}\noindent\textbf{#1.}}
\usepackage{xcolor}
\usepackage{amsmath}

\usepackage[utf8]{inputenc}

\usepackage[skins,breakable]{tcolorbox}
\usepackage{fontawesome5}
\usepackage{lipsum}
\usepackage{longtable}
\usepackage{enumitem}
\usepackage{fontawesome5}
\usepackage{tikz}
\usepackage{amsmath} 
\usepackage{fvextra}
\usepackage{tocloft}
\usepackage{etoc}

\DefineVerbatimEnvironment{verbatim}{Verbatim}{breaklines,breakanywhere}

\definecolor{prompt}{RGB}{223, 223, 192}
\definecolor{prompt-frame}{RGB}{137, 137, 90}
\definecolor{prompt2}{RGB}{223, 223, 192}
\definecolor{prompt2-frame}{RGB}{137, 137, 90}
\definecolor{prompt3}{RGB}{212, 238, 179}
\definecolor{prompt3-frame}{RGB}{117, 146, 77}
\definecolor{prompt4}{RGB}{212, 238, 179}
\definecolor{prompt4-frame}{RGB}{117, 146, 77}
\definecolor{prompt5}{RGB}{212, 238, 179}
\definecolor{prompt5-frame}{RGB}{117, 146, 77}

\tcbset{
    prompt_artist/.style={
        enhanced,
        breakable,
        colback=prompt,        
        colframe=prompt-frame,            
        fontupper=\normalsize,               
        boxrule=1pt,                       
        arc=4mm,                           
        left=1mm, right=1mm, top=1mm, bottom=1mm, 
        boxsep=4pt,                        
        before skip=10pt, after skip=10pt, 
        overlay={
            \node[anchor=north west, xshift=4pt, text=white] at (frame.north west) {\faDatabase};
        },
        title={~~~~~~~\textbf{Prompt: VLM Artist}},
        coltitle=white,
        fonttitle=\bfseries\small
    }
}

\tcbset{
    prompt_style_consistency/.style={
        enhanced,
        breakable,
        colback=prompt,        
        colframe=prompt-frame,            
        fontupper=\normalsize,               
        boxrule=1pt,                       
        arc=4mm,                           
        left=1mm, right=1mm, top=1mm, bottom=1mm, 
        boxsep=4pt,                        
        before skip=10pt, after skip=10pt, 
        overlay={
            \node[anchor=north west, xshift=4pt, text=white] at (frame.north west) {\faDatabase};
        },
        title={~~~~~~~\textbf{Prompt: VLM Artist}},
        coltitle=white,
        fonttitle=\bfseries\small
    }
}

\tcbset{
    prompt_ses_1/.style={
        enhanced,
        breakable,
        colback=prompt,        
        colframe=prompt-frame,            
        fontupper=\normalsize,               
        boxrule=1pt,                       
        arc=4mm,                           
        left=1mm, right=1mm, top=1mm, bottom=1mm, 
        boxsep=4pt,                        
        before skip=10pt, after skip=10pt, 
        overlay={
            \node[anchor=north west, xshift=4pt, text=white] at (frame.north west) {\faDatabase};
        },
        title={~~~~~~~\textbf{Prompt: Initial Sequential Expert System Query}},
        coltitle=white,
        fonttitle=\bfseries\small
    }
}

\tcbset{
    prompt_dataset/.style={
        enhanced,
        breakable,
        colback=prompt,        
        colframe=prompt-frame,            
        fontupper=\normalsize,               
        boxrule=1pt,                       
        arc=4mm,                           
        left=1mm, right=1mm, top=1mm, bottom=1mm, 
        boxsep=4pt,                        
        before skip=10pt, after skip=10pt, 
        overlay={
            \node[anchor=north west, xshift=4pt, text=white] at (frame.north west) {\faDatabase};
        },
        title={~~~~~~~\textbf{Prompt: Dataset Creation}},
        coltitle=white,
        fonttitle=\bfseries\small
    }
}

\tcbset{
    prompt_ses_2/.style={
        enhanced,
        breakable,
        colback=prompt,        
        colframe=prompt-frame,            
        fontupper=\normalsize,               
        boxrule=1pt,                       
        arc=4mm,                           
        left=1mm, right=1mm, top=1mm, bottom=1mm, 
        boxsep=4pt,                        
        before skip=10pt, after skip=10pt, 
        overlay={
            \node[anchor=north west, xshift=4pt, text=white] at (frame.north west) {\faDatabase};
        },
        title={~~~~~~~\textbf{Prompt: Outfit-level negative feedback}},
        coltitle=white,
        fonttitle=\bfseries\small
    }
}

\tcbset{
    negative_flux/.style={
        enhanced,
        breakable,
        colback=prompt,        
        colframe=prompt-frame,            
        fontupper=\normalsize,               
        boxrule=1pt,                       
        arc=4mm,                           
        left=1mm, right=1mm, top=1mm, bottom=1mm, 
        boxsep=4pt,                        
        before skip=10pt, after skip=10pt, 
        overlay={
            \node[anchor=north west, xshift=4pt, text=white] at (frame.north west) {\faDatabase};
        },
        title={~~~~~~~\textbf{Prompt: Try-on-level negative feedback}},
        coltitle=white,
        fonttitle=\bfseries\small
    }
}

\tcbset{
    prompt_negative_search_engine/.style={
        enhanced,
        breakable,
        colback=prompt,        
        colframe=prompt-frame,            
        fontupper=\normalsize,               
        boxrule=1pt,                       
        arc=4mm,                           
        left=1mm, right=1mm, top=1mm, bottom=1mm, 
        boxsep=4pt,                        
        before skip=10pt, after skip=10pt, 
        overlay={
            \node[anchor=north west, xshift=4pt, text=white] at (frame.north west) {\faDatabase};
        },
        title={~~~~~~~\textbf{Prompt: Item-level negative feedback}},
        coltitle=white,
        fonttitle=\bfseries\small
    }
}

\tcbset{
    flux_garment_with_model/.style={
        enhanced,
        breakable,
        colback=prompt,        
        colframe=prompt-frame,            
        fontupper=\normalsize,               
        boxrule=1pt,                       
        arc=4mm,                           
        left=1mm, right=1mm, top=1mm, bottom=1mm, 
        boxsep=4pt,                        
        before skip=10pt, after skip=10pt, 
        overlay={
            \node[anchor=north west, xshift=4pt, text=white] at (frame.north west) {\faDatabase};
        },
        title={~~~~~~~\textbf{Prompt: Garment with figure for virtual-try-on model}},
        coltitle=white,
        fonttitle=\bfseries\small
    }
}

\tcbset{
    flux_garment_without_model/.style={
        enhanced,
        breakable,
        colback=prompt,        
        colframe=prompt-frame,            
        fontupper=\normalsize,               
        boxrule=1pt,                       
        arc=4mm,                           
        left=1mm, right=1mm, top=1mm, bottom=1mm, 
        boxsep=4pt,                        
        before skip=10pt, after skip=10pt, 
        overlay={
            \node[anchor=north west, xshift=4pt, text=white] at (frame.north west) {\faDatabase};
        },
        title={~~~~~~~\textbf{Prompt: Garment without figure for virtual-try-on model}},
        coltitle=white,
        fonttitle=\bfseries\small
    }
}

\title{\name{}: Towards Personalized Fashion Styling via Hierarchical Negative Feedback}

\author{
 Hongbo Ma \\
  Tsinghua University\\
   \And
 Fei Shen \\
   National University of Singapore\\
  \And
 Hongbin Xu \\
  ByteDance Seed\\
  \AND
  Xiaoce Wang \\
  Tsinghua University
  \And
  Gang Xu \\
  Guangming Laboratory \\
  \And
  Jinkai Zheng \\
  Hangzhou Dianzi University \\
  \And
  Liangqiong Qu \\
  University of Hong Kong \\
  \And
  Ming Li \thanks{Corresponding to: liming@gml.ac.cn}\\
  Guangming Laboratory \\
}

\begin{document}
\maketitle
\begin{abstract}
The advancement of intelligent agents has revolutionized problem-solving across diverse domains, yet solutions for personalized fashion styling remain underexplored, which holds immense promise for promoting shopping experiences. In this work, we present \name{}, the first collaborative agent framework that seamlessly unifies personalized apparel design, shopping recommendation, virtual try-on, and systematic evaluation into a cohesive workflow. To this end, \name{} pioneers an iterative visual refinement paradigm driven by \textit{multi-level negative feedback}, enabling adaptive and precise user alignment. 
Specifically, our framework features two core agents, \ie, \textit{Designer} for personalized garment selection and \textit{Consultant} for virtual try-on, whose outputs are progressively refined via hierarchical vision-language model feedback spanning individual items, complete outfits, and try-on efficacy. Counterexamples are aggregated into negative prompts, forming a closed-loop mechanism that enhances recommendation quality.
To assess the performance, we introduce a \textit{comprehensive evaluation suite} encompassing style consistency, visual quality, face similarity, and artistic appraisal. Extensive experiments demonstrate \name{}'s superior performance in delivering personalized designs and recommendations, outperforming strong baselines without negative feedback and establishing a new benchmark for intelligent fashion systems. 
\end{abstract}


\section{Introduction}

Multimodal learning \cite{gpt3,llama,agentai,qwen} has achieved remarkable breakthroughs in recent years, enabling models to seamlessly integrate and process diverse data modalities, thereby transforming their role in practical applications. Intelligent agents, built upon these advancements, represent a powerful paradigm for creating automatic workflows for complicated and originally human-interactive tasks \cite{matchat,paper2code,paper2poster,chartgeneration,foamagent}. 
For example, agents have significantly streamlined paperwork-related activities by enabling automatic chart generation \cite{chartgeneration}, paper-to-code conversion \cite{paper2code}, and the creation of scientific posters from manuscripts \cite{paper2poster}. Beyond administrative automation, agents are also transforming scientific disciplines that demand specialized expertise, such as computational fluid dynamics simulation \cite{matchat,foamagent}. These applications underscore the versatility and impact of agent-based systems.

Although substantial advancements have been made, the development of a unified agent framework for personalized fashion styling remains unaddressed. By tailoring recommendations to user preferences and appearance characteristics, personalized styling systems streamline decision-making, elevate user satisfaction, and are pivotal for increasing e-commerce traffic and operational efficiency~\cite{tryondiffusion, shen2025imagdressing, shen2025imaggarment}.

\begin{figure}[t]
\centering
\includegraphics[width=0.95\columnwidth]{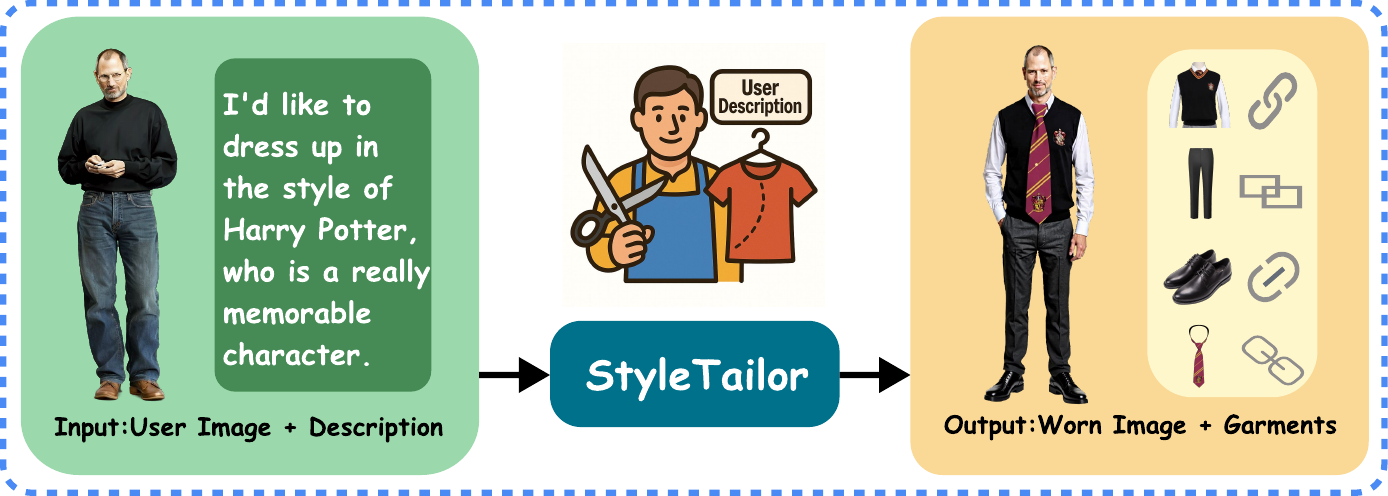} 

\vspace{-2mm}
\caption{We present \name{}, the \textit{first} agentic framework for personalized fashion styling that, given a full-body photo and dressing preferences, outputs virtual try-on results, curated garment images, and shopping links in a unified, closed-loop pipeline, advancing user-centric, interactive fashion recommendation.}
\vspace{-4mm}
\label{fig1}
\end{figure}

However, constructing a collaborative agent framework for personalized fashion styling presents considerable technical difficulties owing to the intricate, fine-grained nature of the task. Despite significant progress in vision-language models (VLMs), existing systems continue to suffer from limited reasoning capabilities and persistent hallucination effects. Disparities in training data distributions and architectural differences among various VLMs further exacerbate inconsistencies, resulting in unreliable and highly domain-dependent outputs \cite{hallucination_survey_1,diversity_data,hallucination_survey_2}. Consequently, ensuring robust reliability, accuracy, and trustworthiness in VLM-based applications remains a substantial challenge.

Accordingly, the effective coordination of multiple agents to exploit the complementary advantages of heterogeneous VLMs constitutes a critical and unresolved research problem. Prevailing agentic frameworks predominantly rely on simplistic, repetitive selection or random output refinement strategies, which incur substantial computational cost yet often yield limited qualitative improvement \cite{autoagent}. Thus, establishing principled mechanisms for efficient feedback integration is an essential challenge for advancing collaborative agent systems in complex, user-centric scenarios.

To address these challenges, we introduce \name{}, the first collaborative agent framework with a \textit{hierarchical negative feedback} mechanism to seamlessly integrate individualized garment design, shopping recommendation, virtual try-on, and systematic evaluation within a unified pipeline. The proposed framework is composed of two principal modules, \ie, \textit{Designer} and \textit{Consultant}. As illustrated in Fig.~\ref{fig1}, \name{} receives a user-provided reference image and dressing style preference description, and outputs virtual try-on visualization, associated garment visuals, and direct purchase links, providing an end-to-end solution for personalized fashion experiences.

The \textit{Designer} employs a cascade of sequential expert agents, each interpreting the inputs by a VLM to generate a standardized set of fine-grained garment specifications across clothing components. Leveraging these attributes, a search engine (\eg, Google Custom Search API) retrieves curated garment images and associated product links. To enforce text-outfit consistency and enable iterative refinement, we incorporate a two-level negative feedback mechanism: (i) at the item level, during the search phase, a VLM analyzes discrepancies between unsatisfactory results and the original prompt, converting them into negative prompts to guide subsequent searches; and (ii) at the outfit level, where the current expert proposes a complete outfit set—if deemed unsatisfactory, the next expert is activated, incorporating prior suboptimal outputs as explicit negative examples, continuing until convergence on a high-quality result.

The \textit{Consultant} utilizes an advanced image-editing model to enable virtual try-on, synthesizing photorealistic images of the user in recommended outfits conditioned on both visual inputs, \ie, user and garment images, and textual prompts. To achieve precise fashion evaluation and alignment with user preferences, we introduce a higher-level negative feedback paradigm that iteratively refines try-on visualizations. The suboptimal results are scrutinized by a VLM to identify discrepancies, which are then converted into negative prompts guiding subsequent generations until optimal consistency and quality are attained.

To comprehensively evaluate our effectiveness, we propose an assessment suite comprising complementary metrics tailored to personalized fashion styling. Style consistency is quantified via VQAScore \cite{vqa}, assessing alignment between synthesized images and user preferences. Visual quality is evaluated using IQAScore \cite{iqa1}, verifying high-fidelity generative outputs. Face similarity is measured with InsightFace \cite{face1}, ensuring minimal identity distortion. Finally, aesthetic appraisal leverages VLM-based evaluators for a holistic artistic and stylistic critique. This suite establishes a robust benchmark for agent-driven fashion systems, enabling precise validation of refinement mechanisms.

Our main contributions can be summarized as follows.
\begin{itemize}
\item We introduce \name{}, the first collaborative agent framework that seamlessly integrates personalized fashion design, shopping recommendation, virtual try-on, and systematic evaluation into a unified pipeline, addressing a key gap in multimodal computer vision for user-centric applications.
\item We propose a hierarchical negative feedback mechanism embedded within the agent system, spanning three progressive levels: item-specific refinement, outfit-level coordination, and virtual try-on optimization. This iterative approach leverages vision-language models to enhance accuracy, mitigate hallucinations, and enable adaptive visual refinement.
\item We present a comprehensive evaluation suite, assessing style consistency, visual quality, facial similarity, and holistic aesthetic appraisal, thereby establishing a robust benchmark for agent-driven fashion systems.
\end{itemize}

\section{Related Works}

\subsection{Multimodal Agents}

The recent surge in intelligent agents has demonstrated their versatility across domains such as software engineering, scientific computing, and visual content creation. These agents, often powered by large language models (LLMs), excel at decomposing complex tasks into actionable steps, enabling sequential reasoning, dynamic interaction, and multi-modal generation~\cite{react, instructllm}. For example, agents have been developed for generating slides and posters~\cite{pptagent, paper2poster}, creating scientific charts~\cite{chartgeneration}, writing code for research or software development~\cite{softwareagent, paper2code}, and supporting scientific simulations in physics and chemistry~\cite{matchat, foamagent}. These applications highlight the powerful generalization and coordination capabilities of LLM-driven agents.
Despite these advances, little attention has been given to agent-based solutions for personalized fashion workflows, which require coordinated garment design, recommendation, and virtual try-on. Moreover, existing agents often lack explicit mechanisms for iterative refinement and user feedback integration, limiting their adaptability in fashion-related tasks.

\begin{figure*}[t]
\centering
\includegraphics[width=0.95\textwidth]{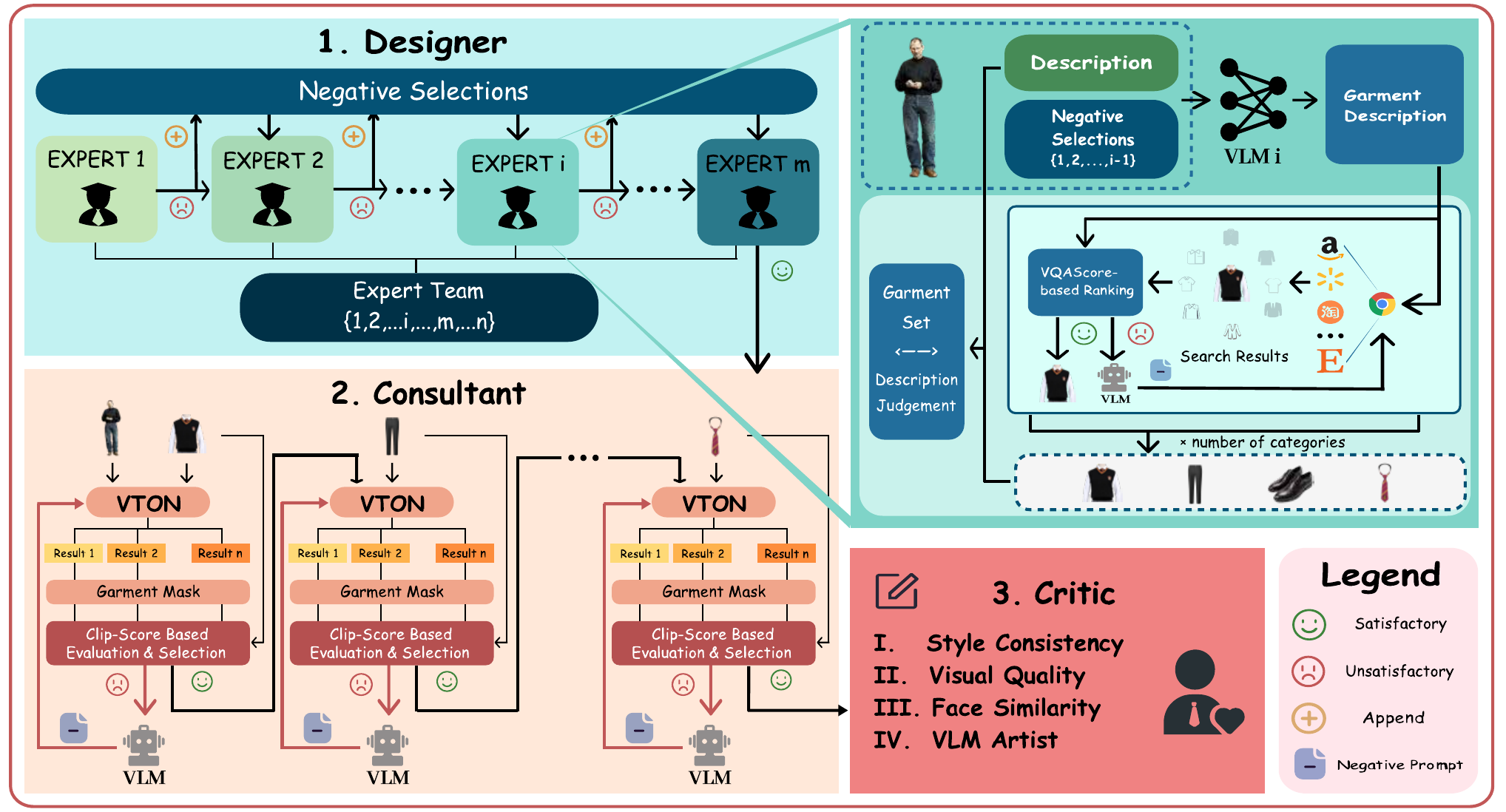} 
\vspace{-2mm}
\caption{Overview of our agent framework \name{}. The \textit{Designer} module analyzes the user-provided image and style preferences, generates garment specifications, and retrieves suitable clothing images. Two hierarchical negative feedback mechanisms within the \textit{Designer} refine retrieval at the item and outfit levels. The \textit{Consultant} module generates virtual try-on results and applies higher-level feedback to further improve alignment with user requirements. The \textit{Critic} quantitatively evaluates the final outputs. This multi-stage feedback ensures the system progressively optimizes recommendations.}
\label{fig_overview}
\vspace{-3mm}
\end{figure*}

\subsection{Fashion Virtual Try-on Models}
Virtual try-on (VTON) methods can be broadly categorized into two paradigms based on input modality: image-based and text/prompt-based.
Image-based models receive a user image and a reference garment to generate try-on results. Early approaches such as VITON~\cite{viton_gan} and CP-VTON introduced warping and composition networks under GAN frameworks, followed by flow-based~\cite{flow_vton} and 3D-aware models~\cite{3dvton} that improved alignment and realism. Recently, diffusion-based methods~\cite{tryondiffusion, shen2025imagdressing, shen2025imaggarment} have achieved high-fidelity synthesis with progressive denoising, while others enhance flexibility by handling pose variation~\cite{swapnet} or relaxing paired-data constraints~\cite{unpaired_data}.
Prompt-based models use textual or multimodal prompts to guide clothing manipulation. Text2Human~\cite{text2human} formulates the task as text-to-image synthesis, while multimodal systems~\cite{multimodal} combine text and reference images for joint control. With the rise of diffusion frameworks such as Stable Diffusion~\cite{stablediffusion} and ControlNet~\cite{controlnet}, text-guided editing has become more precise and scalable.
In this work, we adopt FLUX.1.Kontext~\cite{flux1kontext} as the core image-editing model for virtual try-on. This powerful framework supports joint conditioning on both visual and textual modalities, enabling fine-grained control over garment rendering and accessory augmentation through seamless multimodal guidance.

\section{\name{}}

\subsection{Problem Formulation}
We define personalized fashion styling as a collaborative generation and selection task. Given a full-body reference image of the user \(I_0\) and a natural language description of their dressing preference \(P\), the objective is to generate a realistic try-on image \(I_K\) of the user dressed in a complete outfit, along with the corresponding garment images \(\{G_i\}_{i=1}^{K}\) and their shopping links \(\{L_i\}_{i=1}^{K}\). This task involves multiple challenges, including accurate interpretation of the multimodal input \((I_0, P)\), retrieval of fine-grained garments \(G_i\) that align with the described style, composition of coherent outfits across different clothing components, and photorealistic synthesis of the try-on result \(I_K\) that preserves user identity. To address these challenges, our framework \(\mathcal{T}\) adopts an iterative refinement strategy driven by vision-language feedback, progressively optimizing garment selection and try-on quality to ensure alignment with user intent and enhance the overall personalization experience.

\subsection{Framework Overview}
As illustrated in Fig.~\ref{fig_overview}, StyleTailor consists of two core agents: \textit{Designer} and \textit{Consultant}, which collaboratively drive the personalized fashion styling workflow. The \textit{Designer} module, denoted as \(\mathcal{T}_1\), receives the user’s reference image \(I_0\) and their style preference description \(P\). It then retrieves a set of garment images \(\{G_i\}_{i=1}^{K}\) along with corresponding shopping links \(\{L_i\}_{i=1}^{K}\). These garments are selected to align with the user’s intended aesthetic and are ready for visualization or purchase. The \textit{Consultant} module, denoted as \(\mathcal{T}_2\), takes the initial user image \(I_0\) and the retrieved garments \(\{G_i\}_{i=1}^{K}\) as input and synthesizes a photorealistic try-on image \(I_K\), allowing the user to preview the recommended outfit on themselves. Although the \textit{Designer} and \textit{Consultant} operate independently, their integration forms a unified pipeline where the output of the \textit{Designer} naturally serves as input to the \textit{Consultant}. This modular yet cohesive design enables personalized, end-to-end fashion recommendation and virtual try-on in a structured and adaptive manner.

\subsection{Designer} 

In fashion shopping, a good advisor translates abstract user preferences into specific garment selections. Inspired by this, our \textit{Designer} agent serves as a virtual stylist that interprets user intent and retrieves suitable clothing items from real-world sources. To enhance accuracy and flexibility, we construct a pool of Design Experts, each comprising a \emph{Style Interpreter} and a \emph{Shopping Advisor}, coordinated by a sequential mechanism.

\bfsection{Style Interpreter}  
Users often express dressing intentions in vague or imaginative terms, rather than specific clothing attributes. The Style Interpreter leverages a vision-language model (VLM) to transform the user’s input into a structured representation of garment components. Let \(I_0\) denote the user’s full-body reference image and \(P\) the textual description of their desired style. The VLM used by the \(i\)-th expert is denoted as \(V_i\), and two prompt templates \(t_1\) and \(t_2\) are designed for first-time generation and feedback-refined iterations, respectively. For the first expert (\(i = 1\)), only the raw input \((I_0, P)\) is provided, while for subsequent experts (\(i > 1\)), we also include a negative selection set \(\{s_j\}_{j = 1}^{i - 1}\) representing previously rejected results. Each expert generates a set of category-description pairs \(\{c_i, d_i\}_{i = 1}^{K}\), where \(c_i\) refers to the name of a garment component (e.g., "dresses") and \(d_i\) its professional-level description:
{\small
\begin{equation}
\{c_i, d_i\}_{i = 1}^{K} = 
\begin{cases}
V_1(I_0, P, t_1) & \text{if } i = 1\\
V_i(I_0, P, t_2, \{s_j\}_{j = 1}^{i - 1}) & \text{otherwise}.
\end{cases}
\label{eq:category_generation}
\end{equation}}The interpreter also produces concise summaries for each component, used by the downstream virtual try-on module.

\bfsection{Shopping Advisor}  
Given the structured garment descriptions, the Shopping Advisor performs web-based retrieval to simulate an online shopping process. Let \(d\) denote the textual description of a garment component. A custom search engine \(\mathcal{G}\), built on Google Custom Search API, returns a set of candidate garment images \(\{g_i\}_{i=1}^{M}\) and corresponding shopping links \(\{l_i\}_{i=1}^{M}\):
{\small
\begin{equation}
\{g_i, l_i\}_{i=1}^{M} = \mathcal{G}(d),
\label{eq:garment_search}
\end{equation}}Each candidate is evaluated using VQAScore, which measures the visual-semantic consistency between the retrieved image \(g_i\) and the intended description \(d\). The best-matching image is selected as:
{\small
\begin{equation}
g_0 = \arg\max_{1 \leq i \leq M} \text{VQAScore}(g_i, d).
\label{eq:select_best_image}
\end{equation}}We introduce an \textit{item-level} feedback to iteratively refine the searching results. If the highest score exceeds a predefined threshold \(\tau\), the result is accepted. Otherwise, we use a VLM to analyze mismatches between \(g_0\) and \(d\), extract undesirable features as negative cues, and rerun the search with negation terms. This process is repeated until a satisfactory result is found or a preset iteration limit is reached.

While the Shopping Advisor ensures alignment for individual items, it cannot guarantee global coherence across the outfit. We address this by implementing an \textit{outfit-level} feedback to assess the quality of each full outfit using VQAScore. For each generated garment \(G_i\) and its corresponding component description \(d_i\), we compute a raw alignment score \(s_i = \text{VQAScore}(G_i, P)\), then normalize it by the garment-specific threshold \(\tau_i\) to obtain a bounded score \(s_i' = \min(s_i / \tau_i, 1)\). Assuming independence across components, we aggregate scores using the geometric average:
{\small
\begin{equation}
s_0 = \left( \prod_{i=1}^{K} s_i' \right)^{\frac{1}{K}}.
\label{eq:expert_score}
\end{equation}}If the final score \(s_0\) exceeds an acceptance threshold \(\omega\), the result is accepted. Otherwise, the next expert in the ranked pool is invoked, using the failed attempts as additional negative context. Experts are ordered by their VLM capability (e.g., leaderboard performance), enabling a greedy strategy that balances accuracy and computational efficiency. This multi-level feedback ensures that the generated garments are not only locally relevant but also globally consistent with the user's style intent.

\subsection{Consultant}

After shopping, trying on clothes is a crucial step in determining whether the selected garments truly align with the user's needs and expectations. The \textit{Consultant} module is designed to simulate this virtual try-on process, enabling systematic visual evaluation of the outfit generated by the \textit{Designer}.

\bfsection{Progressive Try-on Generation} 
Let \(I_0\) denote the original user image and \(\{G_i\}_{i = 1}^{K}\) the set of selected garment images. The \textit{Consultant} module \(\mathcal{F}\) generates the final try-on result \(I_K\) by sequentially replacing each garment through \(K\) independent sub-processes \(\{\mathcal{F}_i\}_{i = 1}^K\), each responsible for updating one clothing component:
{\small
\begin{equation}
\mathcal{F} = \mathcal{F}_1 \circ \mathcal{F}_2 \circ \cdots \circ \mathcal{F}_K.
\label{eq:tryon_pipeline}
\end{equation}}
Each sub-process \(\mathcal{F}_i\) receives the image \(I_{i-1}\) from the previous step and produces the updated output \(I_i\):
{\small
\begin{equation}
I_i = \mathcal{F}_i(I_{i-1}) \quad (1 \leq i \leq K).
\label{eq:stage_forward}
\end{equation}}
To reduce visual interference during editing, garments are sorted in descending order of region size—larger components (e.g., outerwear) are replaced first, followed by smaller ones (e.g., accessories).

\bfsection{VLM-Guided Visual Refinement}
Each sub-process \(\mathcal{F}_i\) is implemented using the image-editing model FLUX.1.Kontext\cite{flux1kontext}, denoted \(\mathcal{K}\). The model takes as input the current user image \(I_{i-1}\), the corresponding garment image \(G_i\), and a textual prompt \(z_i\) summarizing the desired appearance, which is produced by the Style Interpreter of the \textit{Designer}. Let \(\mathcal{P}(z_i)\) represent a prompt formatting function. We concatenate \(I_{i-1}\) and \(G_i\), and generate \(l\) try-on candidates:
{\small
\begin{equation}
\{O_i\}_{i = 1}^{l} = \mathcal{K}(\text{concat}(I_{i-1}, G_i), \mathcal{P}(z_i)).
\label{eq:flux_generation}
\end{equation}}

\begin{figure*}[t]
\centering
\includegraphics[width=0.85\textwidth]{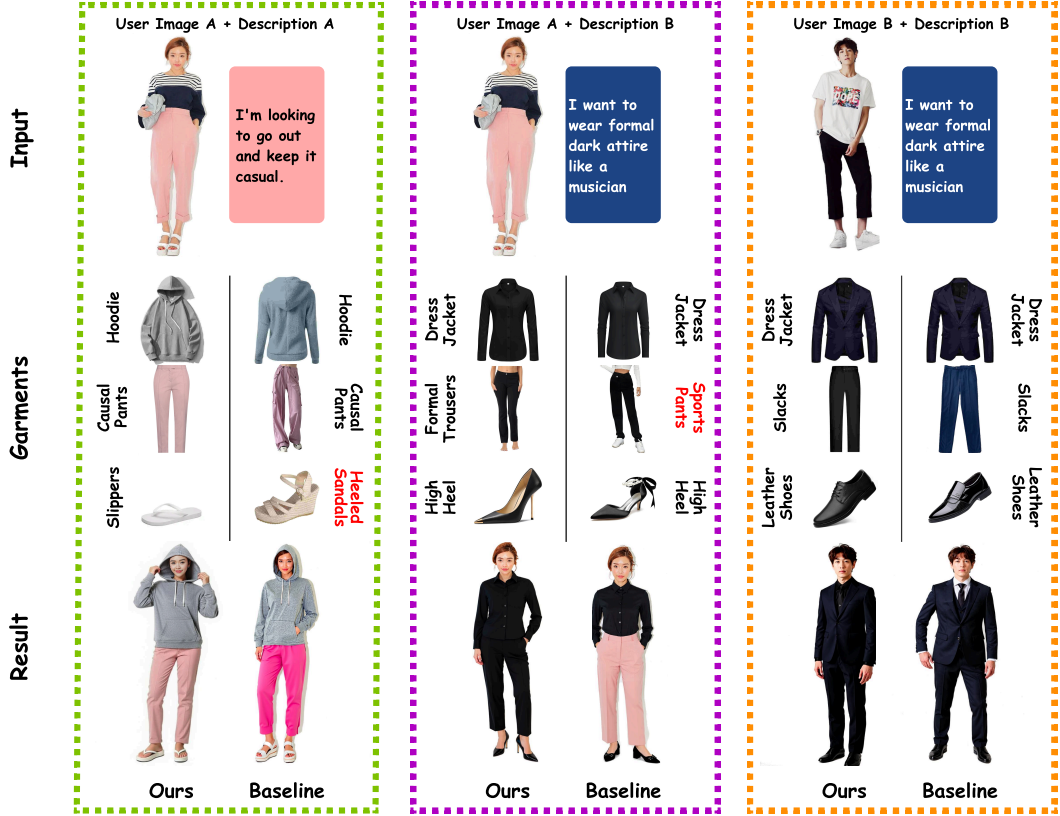} 
\vspace{-2mm}
\caption{Qualitative comparison between our method and the baseline under two conditions: (1) The same user image with different descriptions; (2) The same description with different user images. The visualization results demonstrate both the personalized design capacity of our approach and its superior performance in comparison with the baseline method. The red text indicates the apparently inappropriate garment retrieval by the baseline.}
\label{fig3}
\vspace{-3mm}
\end{figure*}

To select the most accurate result, we apply OpenPose~\cite{openpose1, openpose2, openpose3, openpose4} and HumanParsing~\cite{humanparsing} models to identify the region corresponding to the target category \(c_i\), and extract that region from each generated image. We then compute the CLIPScore~\cite{clipscore} between the masked try-on image and the original garment, and choose the best-matching candidate:
{\small
\begin{equation}
O_0 = \arg\max_{1 \leq i \leq l} \text{CLIPScore}(\text{mask}_{c_i}(O_i), G_i).
\label{eq:clipscore_selection}
\end{equation}}
If the score of \(O_0\) exceeds a predefined threshold \(\sigma\), the result is accepted. Otherwise, the current candidate and garment are passed into a VLM to diagnose visual inconsistencies. These differences are converted into a negative prompt and injected into \(\mathcal{K}\) for regeneration. This \textit{try-on-level} feedback process iterates until the output meets the quality threshold or the maximum number of attempts is reached. The final image \(I_K\) is obtained by applying this process to all garments as defined in Eq.~\eqref{eq:tryon_pipeline}.

\section{Experiments}

\subsection{Settings}

\bfsection{Dataset}
To build the evaluation dataset, we require both the input images and the text prompts.
To effectively balance the data inclusivity with the testing efficiency, we set the dataset size to be 64.
We select input images from an open dataset, LookBook \cite{yoo2016pixel}, which contains a wide variety of high-quality images showcasing different models in various garments.
Subsequently, the text prompts are created through an LLM using the few-shot method \cite{gpt3}.
The text prompts include both specific and abstract requests for suits, dresses, shoes, and other minor accessories.
The composed dataset consists of images of 32 male and 32 female models.
Furthermore, it can be partitioned for each gender according to the criteria mentioned below to enhance diversity:
\begin{itemize}
    \item \textbf{Face Status} of the model is classified into two groups: shown and hidden, maintaining a balanced 1:1 ratio for each gender.
    \item \textbf{Body Status} of the model indicates if the complete body is portrayed or not. Full-body and half-body images are distributed in a 3:1 ratio, retaining uniformity across face status types.
\end{itemize}

\bfsection{Baseline} We take the workflow without any negative feedback mechanisms as a strong baseline. This allows us to verify the effectiveness of our proposed feedback strategies while ensuring the functional completeness of our agent.

\begin{figure*}[t]
\centering
\includegraphics[width=0.8\textwidth]{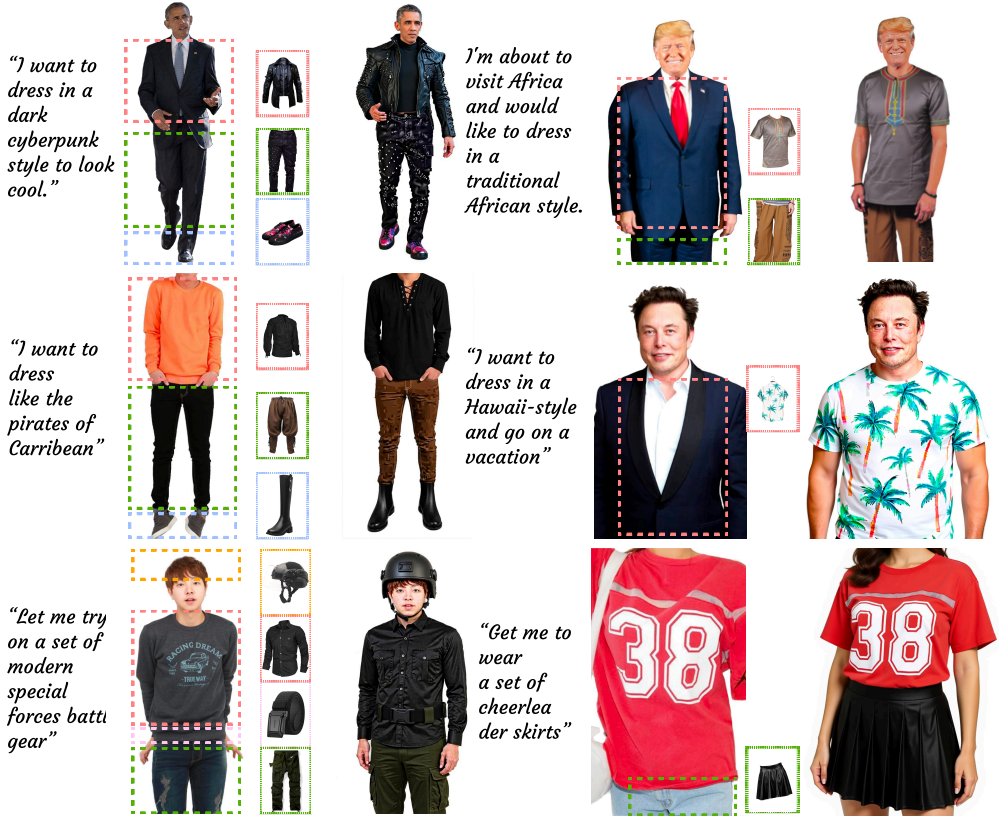} 
\vspace{-2mm}
\caption{Visualizations of diverse user images and style descriptions, along with the corresponding outputs produced by our \name{}. These examples demonstrate \name{}'s ability to effectively handle various user appearances and styling preferences, highlighting its robustness and adaptability to complex, real-world input scenarios.}
\label{fig4}
\vspace{-3mm}
\end{figure*}

\subsection{Evaluation Metrics} 
As illustrated in the \textit{Critic} part of Fig.~\ref{fig_overview}, we apply four metrics to conduct a comprehensive evaluation of the generated image \(I_K\).

\bfsection{Style Consistency} To ensure the alignment between the generated image \(I_K\) and the user's preferences \(P\), we send the user's original image \(I_0\) into a VLM \(V\) to extract human-related attributes excluding the garments, and convert them into text descriptions. We then concatenate these descriptions with the user's initial preferences and input the combined text along with the generated image to compute \(VQAScore(I_K, (V(I_0) + P))\).

\bfsection{Visual Quality} 
To evaluate the generation quality of the generative models, we apply IQAScore\cite{iqa1, iqa2, iqa3} to assess the final generative quality of the output images.

\bfsection{Face Similarity} 
The human face constitutes the most critical feature for personal identification.
To measure the difference between the faces of the original image \(I_0\) and the generated image \(I_K\), we use a pre-trained model from InsightFace\cite{face1} to extract facial features and calculate the cosine similarity to reveal the facial similarity.

\begin{table*}[t]
\centering
\resizebox{0.7\textwidth}{!}{
\begin{tabular}{c|c|c|c|c}
\toprule
Models & Style Consistency & Visual Quality & Face Similarity & VLM Artist \\ \midrule
Baseline & 0.650 & 0.699 & 0.362 & 7.35 \\ 
w/o \textit{Item-level Negative Feedback} & 0.697 & \textbf{0.767} & 0.484 & \underline{8.41}\\
w/o \textit{Outfit-level Negative Feedback} & 0.688 & 0.758 & \underline{0.532} & 8.16 \\
w/o \textit{Try-on-level Negative Feedback} & \underline{0.781} & 0.708 & 0.351 &  8.22 \\
Ours & \textbf{0.906} & \underline{0.764} & \textbf{0.544} & \textbf{8.60} \\ \bottomrule
\end{tabular}
} 
\vspace{-2mm}
\caption{Quantitative comparison. \textbf{Bold} indicates the \textbf{best performance}, \underline{underline} indicates the \underline{second-best performance}.}

\label{table_comparison}
\vspace{-3mm}
\end{table*}

\bfsection{VLM Artist} Besides the alignment between the generated images and the user preferences, as well as the image quality of generative models, we further implement a vision-language model (VLM) artist—specifically, a VLM-based evaluation agent—designed to conduct a comprehensive aesthetic assessment of the final garment synthesis results.
This VLM artist is tasked with evaluating four distinct aspects of the image (detailed below), 
to provide both a short cohesive explanation and a fair integer rating from 1 to 10.

\begin{itemize}

\item \textbf{Design Score} encompasses aesthetic evaluations of individual garment pieces, including \textit{cut} (silhouette and tailoring) and \textit{elements} (decorative accents and embellishments).

\item \textbf{Fitness Score} quantifies the degree of fit between the garments and the wearer's physical attributes (e.g., body shape, proportions).
\item \textbf{Coherence Score} measures the compatibility and stylistic consistency across different garment pieces in the ensemble.
\item \textbf{Mood Score} evaluates the overall mood, stylistic identity, and visual impact conveyed by the entire garment set.
\end{itemize}
These scores, ranging from 1 to 10, are assigned according to detailed criteria described in the Appendix, and their mean constitutes the final output of the VLM artist evaluation.

This multi-dimensional evaluation suite comprehensively assesses both the overall quality of the agent’s outputs and their alignment with user preferences.

\subsection{Comparison with State-of-the-art Methods}

\bfsection{Qualitative Comparison}
As shown in Fig.~\ref{fig3}, the experiments on the same individual with different descriptions revealed that our \name{} can align well with each description.
Using the same description for different individuals, \name{} can still provide personalized solutions tailored to the unique characteristics of each person.
Furthermore, by comparing the experiments within each group, we observe that: in the first group, the baseline shows a noticeably weaker alignment between the shoes and the textual description compared to ours; in the second group, the alignment between the pants and the text is also significantly inferior to ours.
It indicates that our item-level and outfit-level negative feedback play a critical role in the garment retrieval process.
Additionally, baseline results had inconsistent pant colors and accessories, and unreasonable changes in facial features and body posture, underscoring the significance of the negative feedback mechanism in virtual try-on.

\bfsection{Quantitative Comparison} As shown in Tab.~\ref{table_comparison}, we found that our \name{} consistently outperforms the baseline across the board through quantitative analysis. The improvement in \textit{Style Consistency} suggests that the item-level and outfit-level negative feedback integrated into the \textit{Designer} module effectively enhances the alignment between textual descriptions and generated images. Furthermore, the gains observed in the \textit{Visual Quality} and \textit{Face Similarity} metrics indicate that the try-on negative feedback within the \textit{Consultant} module contributes by filtering out implausible results during generation. 

\begin{figure}[t]
\centering
\includegraphics[width=0.95\columnwidth]{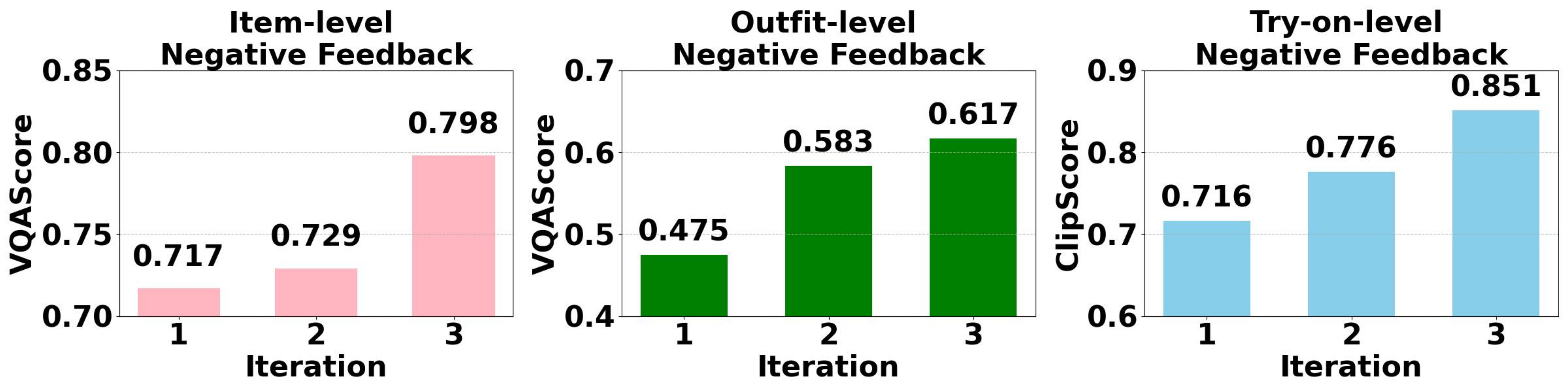} 
\vspace{-2mm}
\caption{The score changes \wrt iterations activated by corresponding negative feedback. The consistent improvements demonstrate the effectiveness of our hierarchical negative feedback mechanism.}
\vspace{-5mm}
\label{fig5}
\end{figure}

The evaluation conducted by the \textit{VLM Artist} reflects an overall assessment of the generated outputs, demonstrating that our method aligns more closely with general aesthetic preferences compared to the baseline. 
On the other hand, we report the corresponding score change tendency of our multi-level negative feedback in Fig. \ref{fig5}. It showcases that the metrics for all negative feedback exhibit a consistent increase over successive iterations. This pattern, when viewed from the standpoint of the execution process, substantiates the efficacy of our negative feedback mechanism.

\subsection{Diversity}
This section aims to demonstrate the adaptability and transferability of our method across complex and diverse tasks. For the user images, we selected images of different genders, various body part presentations, and some celebrities. Regarding users' preferences, we employed descriptions of different styles (\eg, cyberpunk, pirate, vacation, outdoor military). As illustrated in Fig.~\ref{fig4}, our method produced highly adaptive results and performed well across these variations.

\subsection{Ablation Study}
We conduct ablation experiments by respectively removing the proposed negative feedback from our full configurations, as depicted in Tab.~\ref{table_comparison}. We assess the outputs using the metrics proposed above, analyzing the contribution of each negative feedback.

\bfsection{Effect of Item-level Negative Feedback} 
The experiment demonstrates the crucial role of Item-level Negative Feedback in our framework. Removing this component results in a significant decrease in \textit{Style Consistency} (from 0.906 to 0.697) and \textit{VLM Artist} score (from 8.60 to 8.41), confirming that this feedback is essential for accurately aligning garment retrieval with user style preferences.

\bfsection{Effect of Outfit-level Negative Feedback} 
The results of \textit{w/o Outfit-level Negative Feedback} show that the outfit-level feedback is key for ensuring the overall coherence of multi-garment ensembles. Removing this component leads to significant decreases in \textit{Style Consistency} (0.906 to 0.688) and \textit{VLM Artist} score (8.60 to 8.16), indicating that outfit-level feedback is crucial for optimizing the global harmony of recommended outfits, complementing the fine-grained control provided by the item-level feedback.

\bfsection{Effect of Try-on-level Negative Feedback} 
Tab.~\ref{table_comparison} demonstrates that when Try-on-level Negative Feedback is removed, both \textit{Visual Quality} (from 0.764 to 0.708) and \textit{Face Similarity} (from 0.544 to 0.351) experience their most significant declines across all ablations, confirming that high-level feedback at the try-on stage is essential for maintaining clothing fidelity and minimizing unintended facial changes in the generated results. Unlike item- or outfit-level feedback, which primarily refines garment retrieval and global style, this feedback directly supervises the generative model, ensuring both the realism and identity preservation in the final synthesized images. Accurate and realistic virtual try-on results are essential for enabling users to objectively and reliably evaluate how garments and accessories will actually appear when worn, thereby supporting more informed fashion decisions.

\section{Conclusion}

In this paper, we present \textbf{\name{}}, the first agent framework to integrate fashion design, shopping recommendation, and virtual try-on within a unified system. Furthermore, we introduce a multi-level negative feedback mechanism, which enables the agent to continually enhance its reasoning capabilities and iteratively refine its outputs. To ensure comprehensive performance assessment, we propose a set of evaluation metrics tailored to the unique challenges of personalized fashion styling. Extensive experiments demonstrate that \name{} offers substantial application potential in real-world scenarios. The effectiveness and impact of our framework are expected to further expand, opening new directions for intelligent, user-centric fashion and beyond.

\bibliography{main}

\begin{thebibliography}{10}

\bibitem{gpt3}
Tom~B. Brown, Benjamin Mann, Nick Ryder, Melanie Subbiah, Jared Kaplan, Prafulla Dhariwal, Arvind Neelakantan, Pranav Shyam, Girish Sastry, Amanda Askell, Sandhini Agarwal, Ariel Herbert-Voss, Gretchen Krueger, Tom Henighan, Rewon Child, Aditya Ramesh, Daniel~M. Ziegler, Jeffrey Wu, Clemens Winter, Christopher Hesse, Mark Chen, Eric Sigler, Mateusz Litwin, Scott Gray, Benjamin Chess, Jack Clark, Christopher Berner, Sam McCandlish, Alec Radford, Ilya Sutskever, and Dario Amodei.
\newblock Language models are few-shot learners, 2020.

\bibitem{llama}
Hugo Touvron, Thibaut Lavril, Gautier Izacard, Xavier Martinet, Marie-Anne Lachaux, Timothée Lacroix, Baptiste Rozière, Naman Goyal, Eric Hambro, Faisal Azhar, Aurelien Rodriguez, Armand Joulin, Edouard Grave, and Guillaume Lample.
\newblock Llama: Open and efficient foundation language models, 2023.

\bibitem{agentai}
Zane Durante, Qiuyuan Huang, Naoki Wake, Ran Gong, Jae~Sung Park, Bidipta Sarkar, Rohan Taori, Yusuke Noda, Demetri Terzopoulos, Yejin Choi, Katsushi Ikeuchi, Hoi Vo, Li~Fei-Fei, and Jianfeng Gao.
\newblock Agent ai: Surveying the horizons of multimodal interaction, 2024.

\bibitem{qwen}
Qwen, :, An~Yang, Baosong Yang, Beichen Zhang, Binyuan Hui, Bo~Zheng, Bowen Yu, Chengyuan Li, Dayiheng Liu, Fei Huang, Haoran Wei, Huan Lin, Jian Yang, Jianhong Tu, Jianwei Zhang, Jianxin Yang, Jiaxi Yang, Jingren Zhou, Junyang Lin, Kai Dang, Keming Lu, Keqin Bao, Kexin Yang, Le~Yu, Mei Li, Mingfeng Xue, Pei Zhang, Qin Zhu, Rui Men, Runji Lin, Tianhao Li, Tianyi Tang, Tingyu Xia, Xingzhang Ren, Xuancheng Ren, Yang Fan, Yang Su, Yichang Zhang, Yu~Wan, Yuqiong Liu, Zeyu Cui, Zhenru Zhang, and Zihan Qiu.
\newblock Qwen2.5 technical report, 2025.

\bibitem{matchat}
Chen Zi-Yi, Xie Fan-Kai, Wan Meng, Yuan Yang, Liu Miao, Wang Zong-Guo, Meng Sheng, and Wang Yan-Gang.
\newblock Matchat: A large language model and application service platform for materials science.
\newblock {\em Chinese Physics B}, 32(11):118104, November 2023.

\bibitem{paper2code}
Minju Seo, Jinheon Baek, Seongyun Lee, and Sung~Ju Hwang.
\newblock Paper2code: Automating code generation from scientific papers in machine learning, 2025.

\bibitem{paper2poster}
Wei Pang, Kevin~Qinghong Lin, Xiangru Jian, Xi~He, and Philip Torr.
\newblock Paper2poster: Towards multimodal poster automation from scientific papers, 2025.

\bibitem{chartgeneration}
Woosung Koh, Janghan Yoon, MinHyung Lee, Youngjin Song, Jaegwan Cho, Jaehyun Kang, Taehyeon Kim, Se-Young Yun, Youngjae Yu, and Bongshin Lee.
\newblock $c^2$: Scalable auto-feedback for {LLM}-based chart generation.
\newblock In Luis Chiruzzo, Alan Ritter, and Lu~Wang, editors, {\em Proceedings of the 2025 Conference of the Nations of the Americas Chapter of the Association for Computational Linguistics: Human Language Technologies (Volume 1: Long Papers)}, pages 4525--4566, Albuquerque, New Mexico, April 2025. Association for Computational Linguistics.

\bibitem{foamagent}
Ling Yue, Nithin Somasekharan, Yadi Cao, and Shaowu Pan.
\newblock Foam-agent: Towards automated intelligent cfd workflows, 2025.

\bibitem{tryondiffusion}
Luyang Zhu, Dawei Yang, Tyler Zhu, Fitsum Reda, William Chan, Chitwan Saharia, Mohammad Norouzi, and Ira Kemelmacher-Shlizerman.
\newblock Tryondiffusion: A tale of two unets.
\newblock In {\em Proceedings of the IEEE/CVF Conference on Computer Vision and Pattern Recognition (CVPR)}, pages 4606--4615, June 2023.

\bibitem{shen2025imagdressing}
Fei Shen, Xin Jiang, Xin He, Hu~Ye, Cong Wang, Xiaoyu Du, Zechao Li, and Jinhui Tang.
\newblock Imagdressing-v1: Customizable virtual dressing.
\newblock In {\em Proceedings of the AAAI Conference on Artificial Intelligence}, volume~39, pages 6795--6804, 2025.

\bibitem{shen2025imaggarment}
Fei Shen, Jian Yu, Cong Wang, Xin Jiang, Xiaoyu Du, and Jinhui Tang.
\newblock Imaggarment-1: Fine-grained garment generation for controllable fashion design.
\newblock {\em arXiv preprint arXiv:2504.13176}, 2025.

\bibitem{hallucination_survey_1}
Ziwei Ji, Tiezheng Yu, Yan Xu, Nayeon Lee, Etsuko Ishii, and Pascale Fung.
\newblock Towards mitigating hallucination in large language models via self-reflection, 2023.

\bibitem{diversity_data}
Chen Ling, Xujiang Zhao, Jiaying Lu, Chengyuan Deng, Can Zheng, Junxiang Wang, Tanmoy Chowdhury, Yun Li, Hejie Cui, Xuchao Zhang, Tianjiao Zhao, Amit Panalkar, Dhagash Mehta, Stefano Pasquali, Wei Cheng, Haoyu Wang, Yanchi Liu, Zhengzhang Chen, Haifeng Chen, Chris White, Quanquan Gu, Jian Pei, Carl Yang, and Liang Zhao.
\newblock Domain specialization as the key to make large language models disruptive: A comprehensive survey, 2024.

\bibitem{hallucination_survey_2}
Lei Huang, Weijiang Yu, Weitao Ma, Weihong Zhong, Zhangyin Feng, Haotian Wang, Qianglong Chen, Weihua Peng, Xiaocheng Feng, Bing Qin, and Ting Liu.
\newblock A survey on hallucination in large language models: Principles, taxonomy, challenges, and open questions.
\newblock {\em ACM Transactions on Information Systems}, 43(2):1–55, January 2025.

\bibitem{autoagent}
Shuofei Qiao, Ningyu Zhang, Runnan Fang, Yujie Luo, Wangchunshu Zhou, Yuchen~Eleanor Jiang, Chengfei Lv, and Huajun Chen.
\newblock Autoact: Automatic agent learning from scratch for qa via self-planning, 2024.

\bibitem{vqa}
Zhiqiu Lin, Deepak Pathak, Baiqi Li, Jiayao Li, Xide Xia, Graham Neubig, Pengchuan Zhang, and Deva Ramanan.
\newblock Evaluating text-to-visual generation with image-to-text generation.
\newblock {\em arXiv preprint arXiv:2404.01291}, 2024.

\bibitem{iqa1}
Chaofeng Chen and Jiadi Mo.
\newblock {IQA-PyTorch}: Pytorch toolbox for image quality assessment.
\newblock [Online]. Available: \url{https://github.com/chaofengc/IQA-PyTorch}, 2022.

\bibitem{face1}
Xingyu Ren, Alexandros Lattas, Baris Gecer, Jiankang Deng, Chao Ma, and Xiaokang Yang.
\newblock Facial geometric detail recovery via implicit representation.
\newblock In {\em 2023 IEEE 17th International Conference on Automatic Face and Gesture Recognition (FG)}, 2023.

\bibitem{react}
Shunyu Yao, Jeffrey Zhao, Dian Yu, Nan Du, Izhak Shafran, Karthik Narasimhan, and Yuan Cao.
\newblock {ReAct}: Synergizing reasoning and acting in language models.
\newblock In {\em International Conference on Learning Representations (ICLR)}, 2023.

\bibitem{instructllm}
Benfeng Xu, An~Yang, Junyang Lin, Quan Wang, Chang Zhou, Yongdong Zhang, and Zhendong Mao.
\newblock Expertprompting: Instructing large language models to be distinguished experts, 2025.

\bibitem{pptagent}
Hao Zheng, Xinyan Guan, Hao Kong, Jia Zheng, Weixiang Zhou, Hongyu Lin, Yaojie Lu, Ben He, Xianpei Han, and Le~Sun.
\newblock Pptagent: Generating and evaluating presentations beyond text-to-slides, 2025.

\bibitem{softwareagent}
Xingyao Wang, Boxuan Li, Yufan Song, Frank~F. Xu, Xiangru Tang, Mingchen Zhuge, Jiayi Pan, Yueqi Song, Bowen Li, Jaskirat Singh, Hoang~H. Tran, Fuqiang Li, Ren Ma, Mingzhang Zheng, Bill Qian, Yanjun Shao, Niklas Muennighoff, Yizhe Zhang, Binyuan Hui, Junyang Lin, Robert Brennan, Hao Peng, Heng Ji, and Graham Neubig.
\newblock Openhands: An open platform for ai software developers as generalist agents, 2025.

\bibitem{viton_gan}
Xintong Han, Zuxuan Wu, Zhe Wu, Ruichi Yu, and Larry~S. Davis.
\newblock Viton: An image-based virtual try-on network, 2018.

\bibitem{flow_vton}
Xintong Han, Xiaojun Hu, Weilin Huang, and Matthew~R. Scott.
\newblock Clothflow: A flow-based model for clothed person generation.
\newblock In {\em Proceedings of the IEEE/CVF International Conference on Computer Vision (ICCV)}, October 2019.

\bibitem{3dvton}
Shizuma Kubo, Yusuke Iwasawa, Masahiro Suzuki, and Yutaka Matsuo.
\newblock Uvton: Uv mapping to consider the 3d structure of a human in image-based virtual try-on network.
\newblock In {\em Proceedings of the IEEE/CVF International Conference on Computer Vision (ICCV) Workshops}, Oct 2019.

\bibitem{swapnet}
Amit Raj, Patsorn Sangkloy, Huiwen Chang, Jingwan Lu, Duygu Ceylan, and James Hays.
\newblock Swapnet: Garment transfer in single view images.
\newblock In {\em Proceedings of the European Conference on Computer Vision (ECCV)}, September 2018.

\bibitem{unpaired_data}
Assaf Neuberger, Eran Borenstein, Bar Hilleli, Eduard Oks, and Sharon Alpert.
\newblock Image based virtual try-on network from unpaired data.
\newblock In {\em Proceedings of the IEEE/CVF Conference on Computer Vision and Pattern Recognition (CVPR)}, June 2020.

\bibitem{text2human}
Yuming Jiang, Shuai Yang, Haonan Qiu, Wayne Wu, Chen~Change Loy, and Ziwei Liu.
\newblock Text2human: Text-driven controllable human image generation.
\newblock {\em ACM Transactions on Graphics (TOG)}, 41(4):1--11, 2022.

\bibitem{multimodal}
Alberto Baldrati, Davide Morelli, Giuseppe Cartella, Marcella Cornia, Marco Bertini, and Rita Cucchiara.
\newblock Multimodal garment designer: Human-centric latent diffusion models for fashion image editing.
\newblock In {\em Proceedings of the IEEE/CVF International Conference on Computer Vision}, 2023.

\bibitem{stablediffusion}
Robin Rombach, Andreas Blattmann, Dominik Lorenz, Patrick Esser, and Björn Ommer.
\newblock High-resolution image synthesis with latent diffusion models, 2021.

\bibitem{controlnet}
Lvmin Zhang, Anyi Rao, and Maneesh Agrawala.
\newblock Adding conditional control to text-to-image diffusion models, 2023.

\bibitem{flux1kontext}
Black~Forest Labs, Stephen Batifol, Andreas Blattmann, Frederic Boesel, Saksham Consul, Cyril Diagne, Tim Dockhorn, Jack English, Zion English, Patrick Esser, Sumith Kulal, Kyle Lacey, Yam Levi, Cheng Li, Dominik Lorenz, Jonas Müller, Dustin Podell, Robin Rombach, Harry Saini, Axel Sauer, and Luke Smith.
\newblock Flux.1 kontext: Flow matching for in-context image generation and editing in latent space, 2025.

\bibitem{openpose1}
Zhe Cao, Gines Hidalgo, Tomas Simon, Shih-En Wei, and Yaser Sheikh.
\newblock Openpose: Realtime multi-person 2d pose estimation using part affinity fields.
\newblock {\em IEEE Transactions on Pattern Analysis and Machine Intelligence}, 43(1):172--186, 2021.

\bibitem{openpose2}
Zhe Cao, Tomas Simon, Shih-En Wei, and Yaser Sheikh.
\newblock Realtime multi-person 2d pose estimation using part affinity fields, 2017.

\bibitem{openpose3}
Tomas Simon, Hanbyul Joo, Iain Matthews, and Yaser Sheikh.
\newblock Hand keypoint detection in single images using multiview bootstrapping, 2017.

\bibitem{openpose4}
Shih-En Wei, Varun Ramakrishna, Takeo Kanade, and Yaser Sheikh.
\newblock Convolutional pose machines, 2016.

\bibitem{humanparsing}
Peike Li, Yunqiu Xu, Yunchao Wei, and Yi~Yang.
\newblock Self-correction for human parsing, 2019.

\bibitem{clipscore}
Jack Hessel, Ari Holtzman, Maxwell Forbes, Ronan~Le Bras, and Yejin Choi.
\newblock Clipscore: A reference-free evaluation metric for image captioning, 2022.

\bibitem{yoo2016pixel}
Donggeun Yoo, Namil Kim, Sunggyun Park, Anthony~S Paek, and In~So Kweon.
\newblock Pixel-level domain transfer.
\newblock In {\em European conference on computer vision}, pages 517--532. Springer, 2016.

\bibitem{iqa2}
Chaofeng Chen, Jiadi Mo, Jingwen Hou, Haoning Wu, Liang Liao, Wenxiu Sun, Qiong Yan, and Weisi Lin.
\newblock Topiq: A top-down approach from semantics to distortions for image quality assessment.
\newblock {\em IEEE Transactions on Image Processing}, 33:2404--2418, 2024.

\bibitem{iqa3}
Haoning Wu, Zicheng Zhang, Weixia Zhang, Chaofeng Chen, Chunyi Li, Liang Liao, Annan Wang, Erli Zhang, Wenxiu Sun, Qiong Yan, Xiongkuo Min, Guangtai Zhai, and Weisi Lin.
\newblock Q-align: Teaching lmms for visual scoring via discrete text-defined levels.
\newblock {\em International Conference on Machine Learning (ICML)}, 2024.
\newblock Equal Contribution by Wu, Haoning and Zhang, Zicheng. Project Lead by Wu, Haoning. Corresponding Authors: Zhai, Guangtai and Lin, Weisi.

\end{thebibliography}
\bibliographystyle{unsrt}

\clearpage
\appendix
\begin{center}
	\LARGE \bf {Appendix}
\end{center}

\section{Dataset Creation Details}
Dataset creation requires both inclusiveness and practicality. In our elementary experiments, we have noticed that FLUX.1 Kontext, our base model, has a tendency to generate parts of the hidden human body despite not being explicitly asked to. Therefore, a prominent aim of the dataset should be accounting for this phenomenon. To achieve this, we design the dataset to include both images with shown and hidden faces and those with a full body view and with some parts missing.
Regarding the generation process of text input, it is generally a laborious but not a difficult task. After some deliberation, we notice that nearly all requests can be classified into three types:
\begin{itemize}
    \item \textbf{Specific.} Specific prompts call for a specific combination of garments. For example, ``I want to wear a white T-shirt and blue jeans.'' This type of prompt is the most common and the expected usage of our pipeline.
    \item \textbf{Referential.} This type of prompt relies on referencing existing combinations or character designs. For example, ``I want to be like Harry Porter.'' Finding reference is also a common method for people to consider their dress code; therefore, similar requests should be represented in the dataset.
    \item \textbf{Vague.} Though previously improbable to synthesize, due to the ever-enhancing capabilities of base models, our agent should aim to achieve, to a degree, the capability of coping with requests similar to ``I don't know what to wear.'' Therefore, this type of prompt should also be somewhat represented.
\end{itemize}

After deciding on the categories and their respective definitions, it is not difficult to instruct modern LLMs to generate ideal text after prompting the models with a limited set of examples. We performed thorough manual examinations on the generated prompts and have come to the conclusion that the text data is satisfactory.

The specific prompt for instructing the LLM to generate text is as follows:

\begin{tcolorbox}[prompt_dataset]
You are a dataset generator that generates high-quality sample requests for a virtual try-on system. You should output in JSON and follow the following format:
\begin{verbatim}
{
"id": 1, 2, 3...,
"request": YOUR REQUEST
}
\end{verbatim}

\textbf{Template}

\begin{enumerate}
    \item You need to generate requests for certain garments for the virtual try-on system to pick and apply to an image. For example, "I want to wear a white T-shirt" is a valid request.
    \item The data should be comprised of 32 entries for both men and women and include 3 categories, specific, referential and vague.
    \item Specific samples should be the majority, calling for a certain composition of garments. Example: "I want a blue shirt, black pants and a white hat."
    \item Referential samples should refer to an existing character, occupation, etc. Example: "I want to dress like Harry Porter."
    \item Vague samples should be requesting the system to do the decision based on some abstract hints. Example: "Just give me something cute."
\end{enumerate}
Now output your dataset for men and women separately.
\end{tcolorbox}

We have also included the dataset generated using the above method in the supplementary materials.

\section{Implementation Details}

\subsection{Implementation Overview}

The experiments were conducted on a machine with version Ubuntu 22.04, equipped with an NVIDIA RTX 6000 Ada GPU. The clothing categories currently supported in the experiments are: upper body garment, lower body garment, dress, shoes, hat, glasses, belt and, scarf.

\subsection{Designer's Implementation}

\subsubsection{Style Interpreter's Implementation}

In the outfit-level negative feedback, we have prepared four expert systems, each centered on \textit{claude-sonnet-4}, \textit{gemini-2.5-pro}, \textit{llama-4-maverick}, and \textit{qwen-vl-max} in sequence. The threshold $\omega$ for the outfit-level is $0.65$

\subsubsection{Shopping Advisor's Implementation}

During the search engine call, we use the Google Custom Search API, setting the search scope to four online shopping sites—Amazon, Taobao, Walmart, and Etsy and limiting each search to return 10 images. We then set the item-level negative feedback trigger thresholds $\tau$ for the following garment categories: upper body garment, lower body garment, dress, shoes, hat, glasses, belt, and scarf to $[0.7, 0.7, 0.7, 0.6, 0.6, 0.6, 0.6, 0.6]^\top$. What's more, we set the item-level negative feedback mechanism VLM for providing suggestions as \textit{qwen-vl-max}.

\subsection{Consultant's Implementation}

We use FLUX.1.Kontext as the backbone model in virtual try-on and generate 3 images per inference. On the NVIDIA RTX 6000 Ada GPU, each inference takes about 2 minutes. We set the try-on-level negative feedback trigger thresholds $\sigma$ for the following garment categories: upper body garment, lower body garment, dress, shoes, hat, glasses, belt, and scarf to $[0.7, 0.7, 0.7, 0.5, 0.5, 0.6, 0.6, 0.6]^\top$. Furthermore, we set the try-on-level negative feedback mechanism VLM for providing suggestions as \textit{qwen-vl-max}.

\subsection{Critic's Implementation}

In the VLM Artist module, the VLM we use is \textit{qwen-vl-max}.

\section{More Analysis}

\subsection{Efficiency Anaylysis}

\subsubsection{Designer's Efficiency}

We assume that each time we find \textbf{three} garments for the user. Therefore, in the \textit{Designer} module, the runtime of an expert system is the time spent querying the VLM plus the time required for four search engine calls and downloading the corresponding garment images. The time for a single VLM call is about $10$ seconds, and the time for a single search engine call and image downloading is $25$ seconds. Furthermore, in the \textit{Shopping Advisor}, the item-level negative feedback mechanism leads to an additional average of $0.6$ searches, along with $0.6$ additional queries to the VLM. The expert system is called an average of $1.8$ times, so the average runtime of the \textit{Designer} module is:
\[T_{designer} = 1.8 \times [10 + 3 \times (1.6 \times 25 + 0.6 \times 10)] = 266.4 s \approx 4.44 min \]

\subsubsection{Consultant's Efficiency}

We further assume that there are also \textbf{three} garments for the virtual try-on process, with each try-on taking 2 minutes. Due to the outfit-level negative feedback mechanism, the system performs an additional $0.4$ try-ons on average, along with $0.4$ additional queries to the VLM. Thus, the average runtime of the \textit{Consultant} module is:
\[T_{consultant} = 3 \times (1.4 \times 2 \times 60 + 0.4 \times 10) = 515 s \approx 8.6 min\]

\subsubsection{Overall Efficiency}

In the critic module, two additional VLM calls are required, which take $20$ seconds in total ($T_{critic}$). Therefore, the total average time for one round of inference and testing is:

\[T_{overall} = T_{designer} + T_{consultant} + T_{critic} = 266.4 + 515 + 20 = 801.4 s \approx 13.36 min\]

\subsection{Cost Analysis}

The VLMs we use — \textit{claude-sonnet-4}, \textit{gemini-2.5-pro}, \textit{llama-4-maverick}, and \textit{qwen-vl-max} — are all from the OpenRouter platform, and the search engine calls are made using the official Google API. Therefore, our cost analysis is based on the OpenRouter's API pricing and the Google Custom Search API pricing as of August 2025.

\subsubsection{Designer's Cost}

According to the \textbf{Efficiency Analysis} section, assuming \textbf{three} garments are generated, this module requires $8.64$ queries for the search engine, $3.24$ calls to the VLM (\textit{qwen-vl-max}) for the item-level negative feedback and $1.8$ queries to the VLMs (mixture of \textit{claude-sonnet-4}, \textit{gemini-2.5-pro}, \textit{llama-4-maverick}, and \textit{qwen-vl-max}) for the outfit-level negative feedback. When calculating the cost of outfit-level negative feedback, we assign weights $w =(0.4, 0.3, 0.2, 0.1)^\top$ to each component according to the order in which they are called. Therefore, the cost of a single call triggered by the outfit-level negative feedback is:

\begin{align*}
(C'_{\text{input}}, C'_{\text{output}}, C'_{\text{image}}) 
&= \left( \sum_{i = 1}^4 w_i C_{\text{input}_i},\ 
           \sum_{i = 1}^4 w_i C_{\text{output}_i},\ 
           \sum_{i = 1}^4 w_i  C_{\text{image}_i} \right) \\
&= \left( \frac{1.685\ \text{USD}}{M\ \text{tokens}},\ 
          \frac{9.44\ \text{USD}}{M\ \text{tokens}},\ 
          \frac{3.704\ \text{USD}}{K\ \text{imgs}} \right)
\end{align*}

During the item-level negative feedback, $460$ tokens and one image are taken as input and $10$ tokens are taken as output. During the outfit-level negative feedback, $1060$ tokens and one image are taken as input and $40$ tokens are taken as output. Thus the cost of \textit{Designer} module is as follows:

\begin{align*}
C_{\text{designer}} &= 8.64 \times C_{\text{search}} 
+ 3.24 \times \left(460 \times C_{\text{input}} + 10 \times C_{\text{output}} + C_{\text{image}} \right) \\
&\quad + 1.8 \times \left(1060 \times C'_{\text{input}} + 40 \times C'_{\text{output}} + C'_{\text{image}} \right) \\
&= 8.64 \times \frac{5}{1000} + 3.24 \times \left(460 \times \frac{0.8}{1000000} + 10 \times \frac{3.2}{1000000} + \frac{1.024}{1000} \right) \\
&\quad + 1.8 \times \left(1060 \times \frac{1.685}{1000000} + 40 \times \frac{9.44}{1000000} + \frac{3.704}{1000} \right) \\
&= 0.0432 + 0.004614 + 0.010562 = 0.058376 \ \text{USD} \approx 0.058\ \text{USD}
\end{align*}

\subsubsection{Consultant's Cost}

According to the \textbf{Efficiency Analysis} section, assuming \textbf{three} garments are generated, this module requires $1.2$ queries to \textit{qwen-vl-max}. The input per query consists of approximately $500$ tokens and two images, with an output per query of $10$ tokens, so the cost is:
\begin{align*}
C_{\text{consultant}} &= 1.2 \times \left(500 \times C_{\text{input}} + 10 \times C_{\text{output}} + 2 \times C_{\text{image}} \right) \\
&= 1.2 \times \left(500 \times \frac{0.8}{1000000} + 10 \times \frac{3.2}{1000000} + 2 \times \frac{1.024}{1000} \right) \\
&= 0.002976 \text{ USD} \approx 0.003 \text{ USD}
\end{align*}

\subsubsection{Critic's Cost}

In the critic module, two additional VLM calls to \textit{qwen-vl-max} are required. The first VLM for \textit{Style Consistency} requires about $440$ tokens and one image as input and $10$ tokens as output, while the second for \textit{VLM Artist} requires about $780$ tokens and one image as input and $20$ tokens as output. Thus, the cost is:

\begin{align*}
C_{\text{critic}} &= C_1 + C_2 \\ &= (440 + 780) \times C_{\text{input}} + (10 + 20) \times C_{\text{output}} + 2 \times C_{\text{image}} \\
&= 1220 \times \frac{0.8}{1000000} + 30 \times \frac{3.2}{1000000} + 2 \times \frac{1.024}{1000}  \\
&= 0.003088 \text{ USD} \approx 0.003 \text{ USD}
\end{align*}

\subsubsection{Overall Cost}

Thus, the overall cost is the total of the designer's cost, the consultant's cost and the critic's cost.

\begin{align*}
C &= C_{\text{designer}} + C_{\text{consultant}} + C_{\text{critic}} \\ &= 0.058 + 0.003 + 0.003 = 0.064 \text{ USD}
\end{align*}

\section{Limitations and Future Work}

Since this work adopts a training-free approach, the overall performance relies on the capabilities of the pre-defined models within the system. Currently, the limitations of VLM performance are also reflected in our system. However, from another perspective, as VLMs continue to evolve and improve, the performance of our system will also keep advancing. Furthermore, our framework can incorporate additional considerations, such as price, clothing size, and other factors. These can be integrated into user preferences or used as constraints to prune the search process, enabling more powerful and flexible functionalities.

\section{Prompt Templates}

\subsection{Designer's Prompt}

\subsubsection{Style Interpreter's Prompt}

When invoking the first expert, the prompt provided to the VLM within the expert is as follows:

\begin{tcolorbox}[prompt_ses_1]

\textbf{System Prompt:}

You are a clothing prompt generator that converts user clothing requests and their full-body images given to you into detailed text2image prompts for training data generation. You must analyze the user's physical attributes such as gender, ethnics, build, etc. and mix in the user's requests in text to choose suitable clothes for them. You have to follow specific format requirements.

\textbf{Your Task:}
    \begin{enumerate}[label=\arabic*.]
        \item Analyze the user's clothing description and full-body image
        \item \textbf{Your ONLY output must be a single, valid JSON string.}
        \item Generate a detailed prompt in JSON format that describes:
        \begin{itemize}
            \item The requested clothing with rich detail following the template: [style (informal/formal/comic, etc.)], [color],[pattern (dots/stripes, etc.)], [build (slim/wide, etc.)], [texture (wool/silk/artificial, etc.)], [other important details]
            \item Technical specifications for consistent generation
            \item Clothing type classification
            \item You can choose the style from [formal, informal, mode, natural, rock, street, retro, casual, conservative, ethnic, comic]
        \end{itemize}
        \item Extract clothing details including fabric types, fit, color, style category, and decorations
        \item Classify as either "upper body" plus "lower body" or "dresses". Then identify if there are "shoes", "hat", "glasses", "belt" or "scarf" existing.
        \item Output a simple tagged description that could be used to search for the required clothes in online shops. Tags should include descriptions about design details like:
        \begin{itemize}
            \item Color and texture (global features) design
            \item Collar and sleeve (large features) design
            \item Button or zipper (small functional features) design
            \item Style and decorative features
            \item Other valid design tags such as size for hats and heels for shoes for each type of garment you are asked to generate 
        \end{itemize}
        \item Also include a shortened, integrated version of the description required.
    \end{enumerate}

\textbf{Template:}
    \begin{enumerate}[label=\arabic*.]
        \item Analyze the user's clothing description: "{{ user clothing description }}"
        \item Examine the provided full-body image to understand physical attributes
        \item Generate detailed clothing prompts in Mandarin Chinese following the format requirements
        \item Your output \textbf{MUST BE} a single, valid JSON string. Do NOT include any additional text, explanations, or markdown code blocks (e.g., json)
        \item The JSON structure should be exactly as follows:
        \begin{verbatim}
{
 "category": [a list of all valid categories that you see from "upper body", "lower body", "dresses", "shoes", "hat", "glasses", "belt" or "scarf"],
 "prompts": {
    "gender": "[MUST HAVE the person's gender, man / woman]",
    "upper body": "[MUST HAVE tagged upper garment description IN ENGLISH if you see upper and lower clothes else empty]",
    "upper body short": [MUST HAVE shortened version of the description of upper body IN ENGLISH if you see upper + lower clothes else empty],
    "lower body": "[MUST HAVE tagged lower garment description IN ENGLISH if you see upper + lower clothes else empty]",
    "lower body short": [MUST HAVE shortened version of the description of lower body IN ENGLISH if you see upper + lower clothes else empty],
    "dresses": "[MUST HAVE tagged dress description IN ENGLISH if category is dress else empty]",
    "dresses short": [MUST HAVE shortened version of the description of dresses IN ENGLISH if category is dress else empty],
    "shoes": "[tagged shoes description IN ENGLISH if you see shoes else empty]",
    "shoes short": [shortened version of the description of shoes IN ENGLISH if you see shoes else empty],
    "hat": "[tagged hat description IN ENGLISH if you see hat else empty]",
    "hat short": [shortened version of the description of hat IN ENGLISH if you see hat else empty],
    "glasses: "[tagged glasses description IN ENGLISH if you see glasses else empty]",
    "glasses short": [shortened version of the description of glasses IN ENGLISH if you see glasses else empty],
    "belt": "[tagged belt description IN ENGLISH if you see belt else empty]",
    "belt short": [shortened version of the description of belt IN ENGLISH if you see belt else empty],
    "scarf": "[tagged scarf description IN ENGLISH if you see scarf else empty]",
    "scarf short": [shortened version of the description of scarf IN ENGLISH if you see scarf else empty]
 }
}
        \end{verbatim}
        \item Each full length prompt section should include gender preference, type, design, build and other valueable details in short phrases and should end with HD, no model.
        \item \textbf{YOU MUST INCLUDE THE TYPE OF CLOTHES TO SEARCH} like shirt, T-shirt, jeans, pants, dress, skirt or something else.
        \item  Examples are given below:
        \begin{itemize}
            \item For upper body: "Men's, white, 100 percent cotton, mandarin collar, long sleeves, formal shirt, loose cut, chest pocket, HD, no model."
            \item For upper body short: "long-sleeved mandarin collar white shirt with a pocket."
            \item For lower body: "Women's, dark blue, denim, high-waisted, straight-leg jeans, slim cut, distressed detailing, HD, no model."
            \item For lower body short: "blue high-waist long straight-leg jeans with details."
            \item For dresses: "Women's, pink, silk, pleated, slim cut, short tail, chequered decor, HD, no model."
            \item For dresses short: "pink silk pleated dress."
        \end{itemize}
        \item Always ensure valid JSON output
        \item "upper body" + "lower body" or "dresses" and their shortened version are \textbf{MUST HAVES}. Other categories are optional depending on the image
    \end{enumerate}
  User clothing request: {{ user clothing description }}
  
  Now consider the given image of their portrait given alongside their requests.

\textbf{Jinja Args:}

  - user clothing description

\end{tcolorbox}

When the output result of the previous expert is unsatisfactory, the prompt provided to the VLM of the next expert is as follows:

\begin{tcolorbox}[prompt_ses_2]
\textbf{System Prompt:}
You are a clothing prompt generator that converts user clothing requests and their full-body images given to you into detailed text2image prompts for training data generation. You must analyze the user's physical attributes such as gender, ethnics, build, etc. and mix in the user's requests in text to choose suitable clothes for them. You have to follow specific format requirements.

\textbf{Your Task:}
\begin{enumerate}[label=\arabic*.]
    \item Analyze the user's clothing description and full-body image
    \item \textbf{Your ONLY output must be a single, valid JSON string.}
    \item Generate a detailed prompt in JSON format that describes:
        \begin{itemize}
            \item The requested clothing with rich detail following the template: [style (informal/formal/comic, etc.)], [color], [pattern (dots/stripes, etc.)], [build (slim/wide, etc.)], [texture (wool/silk/artificial, etc.)], [other important details]
            \item Technical specifications for consistent generation
            \item Clothing type classification
            \item You can choose the style from [formal, informal, mode, natural, rock, street, retro, casual, conservative, ethnic, comic]
        \end{itemize} 
    \item Extract clothing details including fabric types, fit, color, style category, and decorations
    \item Classify as either "upper body" + "lower body" or "dresses". Then identify if there are "shoes", "hat", "glasses", "belt" or "scarf" existing.
    \item Output a simple tagged description that could be used to search for the required clothes in online shops. Tags should include descriptions about design details like:
        \begin{itemize}
          \item Color and texture (global features) design
          \item Collar and sleeve (large features) design
          \item Button or zipper (small functional features) design
          \item Style and decorative features
          \item Other valid design tags such as size for hats and heels for shoes for each type of garment you are asked to generate 
        \end{itemize} 
    \item Also include a shortened, integrated version of the description required.    
\end{enumerate}

\textbf{Template:}
  Instructions:
  \begin{enumerate}[label=\arabic*.]
      \item Analyze the user's clothing description: "{{ user clothing description }}"
      \item Examine the provided full-body image to understand physical attributes
      \item Generate detailed clothing prompts in Mandarin Chinese following the format requirements
      \item Your output \textbf{MUST BE} a single, valid JSON string. Do NOT include any additional text, explanations, or markdown code blocks (e.g., ```json).
      \item The JSON structure should be exactly as follows:
      \begin{verbatim}
{
"category": [a list of all valid categories that you see from "upper body", "lower body", "dresses", "shoes", "hat", "glasses", "belt" or "scarf"],
"prompts": {
    "gender": "[MUST HAVE the person's gender, man / woman]",
    "upper body": "[MUST HAVE tagged upper garment description IN ENGLISH if you see upper + lower clothes else empty]",
    "upper body short": [MUST HAVE shortened version of the description of upper body IN ENGLISH if you see upper + lower clothes else empty],
    "lower body": "[MUST HAVE tagged lower garment description IN ENGLISH if you see upper + lower clothes else empty]", 
    "lower body short": [MUST HAVE shortened version of the description of lower body IN ENGLISH if you see upper + lower clothes else empty],
    "dresses": "[MUST HAVE tagged dress description IN ENGLISH if category is dress else empty]",
    "dresses short": [MUST HAVE shortened version of the description of dresses IN ENGLISH if category is dress else empty],
    "shoes": "[tagged shoes description IN ENGLISH if you see shoes else empty]",
    "shoes short": [shortened version of the description of shoes IN ENGLISH if you see shoes else empty],
    "hat": "[tagged hat description IN ENGLISH if you see hat else empty]",
    "hat short": [shortened version of the description of hat IN ENGLISH if you see hat else empty],
    "glasses: "[tagged glasses description IN ENGLISH if you see glasses else empty]",
    "glasses short": [shortened version of the description of glasses IN ENGLISH if you see glasses else empty],
    "belt": "[tagged belt description IN ENGLISH if you see belt else empty]",
    "belt short": [shortened version of the description of belt IN ENGLISH if you see belt else empty],
    "scarf": "[tagged scarf description IN ENGLISH if you see scarf else empty]",
    "scarf short": [shortened version of the description of scarf IN ENGLISH if you see scarf else empty]
}
}
      \end{verbatim}
      \item Each full length prompt section should include gender preference, type, design, build and other valueable details in short phrases and should end with HD, no model.
      \item \textbf{YOU MUST INCLUDE THE TYPE OF CLOTHES TO SEARCH} like shirt, T-shirt, jeans, pants, dress, skirt or something else.
      \item Examples are given below:
      \begin{itemize}
          \item For upper body: "Men's, white, 100 percent cotton, mandarin collar, long sleeves, formal shirt, loose cut, chest pocket, HD, no model."
          \item For upper body short: "long-sleeved mandarin collar white shirt with a pocket."
          \item For lower body: "Women's, dark blue, denim, high-waisted, straight-leg jeans, slim cut, distressed detailing, HD, no model."
          \item For lower body short: "blue high-waist long straight-leg jeans with details."
          \item For dresses: "Women's, pink, silk, pleated, slim cut, short tail, chequered decor, HD, no model."
          \item For dresses short: "pink silk pleated dress."
      \end{itemize}
      \item Always ensure valid JSON output
      \item "upper body" + "lower body" or "dresses" and their shortened version are \textbf{MUST HAVES}. Other categories are optional depending on the image
  \end{enumerate} 
  
  User clothing request: {{user clothing description}}
  
  In addition, here are some examples where the user is not satisfied about. Please use them as reference and consider avoiding these answers: {{negative examples}}]
  
  Please think carefully and provide your clothing recommendation.

\textbf{Jinja Args:}

  - user clothing description
  
  - negative examples
    
\end{tcolorbox}

\subsubsection{Shop Advisor's Prompt}

When the retrieved image does not meet the requirements, the prompt provided to the VLM for generating negative feedback on the current image is as follows:

\begin{tcolorbox}[prompt_negative_search_engine]
\textbf{System Prompt}
You are part of a garment search system that analyzes search results to identify issues with retrieved images. Your task is to create abstract negative prompts that prevent similar problems in future searches.

\textbf{Your Responsibilities:}

\begin{enumerate}[label=\arabic*.]
    \item Compare the text description with the retrieved search result image
    \item Identify discrepancies between the requested garment and the actual image
    \item Focus on abstract, generalizable issues rather than specific details
    \item Generate symmetrical positive/negative prompt pairs
    \item Return results in valid JSON format with "positive prompt" and "negative prompt" keys
\end{enumerate}

\textbf{Requirements:}

\begin{enumerate}[label=\arabic*.]
    \item Be abstract and generalizable to any search situation
    \item Focus on fundamental issues like image quality, garment type accuracy, style matching
    \item Avoid overly specific details that won't apply broadly
    \item Ensure positive and negative prompts are complementary opposites
    \item Keep prompts under 3 words each
    \item YOU SHOULD ONLY GENERATE ONE PAIR OF PROMPTS AT A TIME
\end{enumerate}

\textbf{Template:}
  Chain of Thought Process:
  \begin{enumerate}[label=\arabic*.]
      \item Analyze the text description: What garment characteristics were requested?
      \item Examine the search result image: What is actually shown?
      \item Identify the primary discrepancy: Is it about image quality, garment type, style, color, or other attributes?
      \item Abstract the issue: What general category does this problem fall into?
      \item Formulate prompts: Create positive guidance and negative prevention that applies broadly
  \end{enumerate}

\textbf{Example scenarios and outputs:}
    \begin{itemize}
        \item Example 1:
        \begin{verbatim}
Description: "Shoes, Women's, dark brown, leather, high heels, ankle strap, elegant style, HD, no model"
Image: Very low resolution shoes
Issue: Poor image quality
Output:
{
  "positive prompt": ["high resolution"],
  "negative prompt": ["low resolution"]
}
        \end{verbatim}
        \item Example 2:
        \begin{verbatim}
Description: "Men's formal shirt, white, cotton, long sleeve"
Image: Women's blouse
Issue: Wrong gender category
Output:
{
  "positive_prompt": ["correct gender"],
  "negative_prompt": ["wrong gender"]
}
        \end{verbatim}
        \item Example 3:
        \begin{verbatim}
Description: "Black leather jacket, motorcycle style"
Image: Fabric blazer
Issue: Wrong material and style
Output:
{
  "positive prompt": ["correct material"],
  "negative prompt": ["wrong material"]
}    
        \end{verbatim}
    \end{itemize}
    
  Now analyze the provided description and search result image: "{{user clothing description}}"
  
  Follow the CoT process above and provide your response in the specified JSON format that hold lists of prompts.

\textbf{Jinja Args:}

  - user clothing description
    
\end{tcolorbox}

\subsection{Consultant's Prompt}

For garment images containing a human body, the input prompt for FLUX.1.Kontext is as follows

\begin{tcolorbox}[flux_garment_with_model]

\textbf{System Prompt:}
 Make the {{gender}} on the left side wear the {{description}}. Ignore the model wearing the {{description}}. Keep the style and design of the {{description}} the same. Special notice to correct the clothes color after you conduct. Remove the clothes and the right side model from the scene.

\textbf{Jinja Args:}

  - gender
  
  - description
    
\end{tcolorbox}

For garment images without a human body, the input prompt for FLUX.1.Kontext is as follows

\begin{tcolorbox}[flux_garment_without_model]

\textbf{System Prompt:}
Make the {{gender}} wear the {{description}}. Keep the {{gender}}'s pose the same. Keep the style and design of the {{description}} the same. Correct the clothes color after you conduct. Remove the clothes from the scene.

\textbf{Jinja Args:}

  - gender
  
  - description  
  
\end{tcolorbox}

When the post-try-on image does not meet the requirements, the prompt provided to the VLM for generating negative feedback on the current image is as follows:

\begin{tcolorbox}[negative_flux]

\textbf{System Prompt:}
You are part of a virtual try-on agent that analyzes user description and AI generated images to identify issues with AI-generated clothing changes. Your task is to create abstract negative prompts that prevent similar problems in future generations.

\textbf{Your Responsibilities:}
\begin{enumerate}[label=\arabic*.]
    \item Compare the original description with the virtual try-on result
    \item Identify discrepancies between the requested clothing description and the generated result
    \item Focus on abstract, generalizable issues rather than specific details
    \item Generate symmetrical positive/negative prompt pairs
    \item Return results in valid JSON format with "positive prompt" and "negative prompt" keys being list items
\end{enumerate}
  
\textbf{Requirements:}

\begin{itemize}
    \item Be abstract and generalizable to any situation
    \item Focus on fundamental issues like body consistency, garment type accuracy, style matching
    \item Avoid overly specific details that won't apply broadly
    \item Ensure positive and negative prompts are complementary opposites
    \item Keep prompts under 3 words each
\end{itemize}

\textbf{Template:}
Chain of Thought Process:
\begin{enumerate}[label=\arabic*.]
  \item Analyze the original description: What might be the person's key physical characteristics?
  \item Examine the generated result: What clothing was actually produced?
  \item Compare with the requested description: "{{user clothing description}}"
  \item Identify the primary discrepancy: Is it about garment type, fit, style, color, or something else?
  \item Abstract the issue: What general category does this problem fall into?
  \item Formulate prompts: Create positive guidance and negative prevention that applies broadly
\end{enumerate}

\textbf{Example scenarios and outputs:}

\begin{itemize}
    \item Example 1:
    \begin{verbatim}
Description: "women's dress"
Generated: Man in dress
Issue: Gender inconsistency
Output:
{
  "positive_prompt": ["consistent gender"],
  "negative_prompt": ["inconsistent gender"]
}  
    \end{verbatim}
    \item Example 2:
    \begin{verbatim}
Description: "formal blazer"
Generated: Person in casual hoodie
Issue: Formality level mismatch
Output:
{
  "positive_prompt": ["correct formality", "correct style"],
  "negative_prompt": ["wrong formality", "wrong style"]
}  
    \end{verbatim}
    \item Example 3:
    \begin{verbatim}
Description: "blue hat"
Generated: Person in red hat
Issue: Color mismatch
Output:
{
  "positive_prompt": ["blue"],
  "negative_prompt": ["not blue"]
} 
    \end{verbatim}
\end{itemize}

Now analyze the provided image and clothing description: "{{user clothing description}}"
  
Follow the CoT process above and provide your response in the specified JSON format.

Remember to comment on the garment that loosely corresponds to the description. Be generally \textbf{ABSTRACT} but you can also comment on any detailed discrepancies.

\textbf{Jinja Args:}

  - user clothing description
    
\end{tcolorbox}

\subsection{Critic's Prompt}

In this section, the \textbf{Style Consistency} metric requires a VLM to extract the physical appearance features of the user image (excluding garments).

\begin{tcolorbox}[prompt_style_consistency]

\textbf{System Prompt:}
You are a person description generator that analyzes full-body images to extract detailed physical and facial attributes for AI clothing change evaluation metrics.

\textbf{Your task:}

\begin{enumerate}[label=\arabic*.]
    \item Analyze the user's full-body image to extract comprehensive physical attributes
    \item Generate a detailed description that focuses on:
        \begin{itemize}
            \item Facial features and characteristics
            \item Body proportions and physical build
            \item Pose and positioning
            \item Skin tone and hair details
            \item Overall appearance markers for consistency tracking
        \end{itemize}
    \item Ignore all clothing items and accessories
    \item Output detailed physical description in natural English
    \item Maintain 50-100 words for consistent evaluation
    \item Use proper XML tags to enclose description sections
\end{enumerate}

\textbf{Template:}
  Instructions:
  \begin{enumerate}[label=\arabic*.]
      \item Examine the provided full-body image to identify:
      \begin{itemize}
        \item Facial structure: round, oval, square, heart-shaped, angular, etc.
        \item Eye characteristics: size, shape, color if visible, eyebrow shape
        \item Nose features: size, shape, bridge characteristics
        \item Mouth and lip features: size, shape, expression
        \item Hair: color, length, texture, style, hairline
        \item Skin tone: fair, medium, olive, tan, dark, etc.
        \item Body proportions: height indicators, shoulder width, waist-to-hip ratio
        \item Build: slim, athletic, curvy, stocky, muscular, petite, etc.
        \item Pose: standing position, arm placement, leg stance, body angle
        \item Overall physique and distinctive features
      \end{itemize}
      \item Generate comprehensive person description excluding all clothing and accessories
      \item Output in the following format in English:
      \textless person description\textgreater A [ethnicity] [gender] with [facial structure] face shape, [eye description], [nose description], [mouth description], [hair description], [skin tone] complexion, [height build description], [body proportions], [pose description], [distinctive physical features], full body visible, natural pose, clear facial features, consistent person identity\textless/person description\textgreater
      \textless evaluation focus\textgreater [facial/body/pose]\textless/evaluation focus\textgreater

      \item Ensure description is 50-100 words in length
      \item Use \textless evaluation focus \textgreater facial \textless /evaluation focus \textgreater for: primarily face-focused analysis
      \item Use \textless evaluation focus \textgreater body \textless /evaluation focus \textgreater for: body proportion and build focus
      \item Use \textless evaluation focus \textgreater pose \textless /evaluation focus \textgreater for: posture and positioning focus
  \end{enumerate}
\end{tcolorbox}

Besides, the \textbf{VLM Artist} metric requires a VLM for evaluation. The following is its prompt.

\begin{tcolorbox}[prompt_artist]
\textbf{System Prompt:}
You are a fashion and garment aesthetics evaluator that analyzes clothing and styling in images of people wearing garments. Your focus is exclusively on the clothing design, fit, styling, and overall aesthetic coherence.

\textbf{Your Task:}
\begin{enumerate}[label=\arabic*.]
    \item Analyze the garment(s) and styling in the provided image.
    \item Evaluate clothing design elements, fit quality, and aesthetic appeal.
    \item Assess how well the garments work together as a cohesive outfit.
    \item Comment on the mood, style, and visual impact of the clothing.
    \item Never comment on the person's physical appearance, body, or personal attributes.
    \item Focus solely on the garments, their design, fit, and styling choices.
    \item Provide constructive fashion analysis with ratings and comments.
    \item Output your evaluation in valid JSON format.
\end{enumerate}

\textbf{Template:}
\begin{enumerate}[label=\arabic*.]
    \item \textbf{Instructions:}
    \begin{itemize}
        \item Garment design: silhouette, cut, structure, design details
        \item Color palette: harmony, contrast, seasonal appropriateness
        \item Fabric and texture: quality appearance, drape, finish
        \item Fit assessment: how well garments conform to the wearer's form
        \item Style coherence: how pieces work together thematically
        \item Styling choices: layering, proportions, styling techniques
        \item Mood and aesthetic: casual, formal, edgy, romantic, minimalist, etc.
        \item Overall visual impact and fashion-forward appeal
    \end{itemize}
    \item Focus exclusively on clothing and styling - avoid any personal commentary.
    \item Rate each category from 1-10 (1 = poor, 10 = excellent).
    \item Criteria for each category.
    \begin{itemize}
        \item Detailed description of each score for design rating:
    
        1: Impossible to make
        
        2: Very bad design choice
        
        3: Bad design choice such as improper material usage
        
        4: Minor design issue
        
        5: Accpetable but not good
        
        6: Decent design with no mistakes
        
        7: Good design with highlights such as inspiring cut
        
        8: Very good design with innovations
        
        9: Award-winning and impressive design
        
        10: Masterclass in garment designing

        \item Detailed description of each  score for fit rating:

        1: Unwearable
        
        2: Very bad fit
        
        3: Bad fit such as incorrect body ratio
        
        4: Minor fitting issue
        
        5: Accpetable but not good
        
        6: Correct fit with no mistakes
        
        7: Good fit with beautiful wearer presentation
        
        8: Very good fit with innovations
        
        9: Near perfect fit
        
        10: Appears custom tailored with perfect fit

        \item Detailed description of each score for coherence rating:

        1: Not a set at all
        
        2: Very bad coherence
        
        3: Bad coherence such as mismatched formality
        
        4: Minor coherence issue
        
        5: Accpetable but not good
        
        6: Decent coherence with no mistakes
        
        7: Good coherence with the set looking natural and complete
        
        8: Very good coherence with innovations
        
        9: Great set that seems designed as a whole
        
        10: Awesome set that shines on a grand stage

        \item Detailed description of each score for mood rating:

        1: Impossible to understand
        
        2: Very bad mood control
        
        3: Bad garment mood that can not fulfil its purpose
        
        4: Minor mood issue
        
        5: Accpetable but not good
        
        6: Decent mood control with no mistakes
        
        7: Good mood building with  generally memorable visuals
        
        8: Very good mood building
        
        9: Great mood and atmosphere that impacts the eye
        
        10: Awesome mood building that feels real and immersive
        
    \end{itemize}

    \item Output in the following JSON format:
    \begin{verbatim}
{
"design": "Analysis of individual pieces including silhouette, cut, design elements, fabric appearance, and construction quality",
"design rating": 1-10,
"fit": "Evaluation of how well the garments fit the wearer's form, including proportion balance and silhouette enhancement",
"fit rating": 1-10,
"coherence": "Assessment of how the pieces work together as a complete outfit, including color harmony, style consistency, and thematic unity",
"coherence rating": 1-10,
"mood": "Description of the overall mood, style category, and visual impact created by the clothing ensemble",
"mood rating": 1-10,
"overall comment": "Summary of the outfit's strengths and areas for improvement, focusing on the complete aesthetic package",
"overall rating": 1-10
}      
    \end{verbatim}
    \item Ensure comments are concise but informative (50 words each).
    \item Overall rating should reflect the average of individual ratings.
    \item Focus on constructive analysis that evaluates fashion merit.
\end{enumerate}

Below is the image to comment on:

\end{tcolorbox}

\end{document}